# SIMULATION APPROACHES TO GENERAL PROBABILISTIC INFERENCE ON BELIEF NETWORKS

Ross D. Shachter
Department of Engineering-Economic Systems, Stanford University
Terman Engineering Center, Stanford, CA 94305-4025
SHACHTER@SUMEX-AIM.STANFORD.EDU
and
Mark A. Peot
Department of Engineering-Economic Systems, Stanford University
and Rockwell International Science Center, Palo Alto Laboratory
444 High Street, Suite 400, Palo Alto, CA 94301
PEOT@RPAL.COM

Although a number of algorithms have been developed to solve probabilistic inference problems on belief networks, they can be divided into two main groups: exact techniques which exploit the conditional independence revealed when the graph structure is relatively sparse, and probabilistic sampling techniques which exploit the "conductance" of an embedded Markov chain when the conditional probabilities have non-extreme values. In this paper, we investigate a family of Monte Carlo sampling techniques similar to Logic Sampling [Henrion, 1988] which appear to perform well even in some multiply-connected networks with extreme conditional probabilities, and thus would be generally applicable. We consider several enhancements which reduce the posterior variance using this approach and propose a framework and criteria for choosing when to use those enhancements.

## 1. Introduction

Bayesian belief networks or influence diagrams are an increasingly popular representation for reasoning under uncertainty. Although a number of algorithms have been developed to solve probabilistic inference problems on these networks, these prove to be intractable for many practical problems. For example, there are a variety of exact algorithms for general networks, using clique join trees [Lauritzen and Spiegelhalter 1988], conditioning [Pearl 1986b] or arc reversal [Shachter 1986]. All of these algorithms are sensitive to the connectedness of the graph, and even the first, which appears to be the fastest, quickly grows intractable for medium size practical problems. This is not surprising, since the general problem is NP-hard [Cooper 1987]. Alternatively, several Monte Carlo simulation algorithms [Henrion 1988, Pearl 1987, and Chavez 1989] promise polynomial growth in the size of the problem, but suffer from other limitations. Markov chain algorithms such as [Pearl 1987] and [Chavez 1989] may degrade rapidly ($\propto [\ln (1 + p_{min})]^{-1}$) if there are conditional probabilities near zero [Chin and Cooper 1987, Chavez 1989]. Convergence rates for Logic Sampling [Henrion 1988] degrade exponentially with the number of pieces of evidence.

The goal of this research is to develop simulation algorithms which are suitable for a broad range of problem structures, including problems with multiple connectedness, extreme probabilities and even deterministic logical functions. Most likely, these algorithms will not be superior for all problems, but they do seem promising for reasonable general purpose use. In particular, there are several enhancements which can be adaptively applied to improve their performance in a problem-sensitive manner. Best of all, the algorithms described in this paper lend themselves to simple parallel implementation, and can, like nearly all simulation algorithms, be interrupted at "anytime," yielding the best solution available so far.

## 2. The Algorithms

Let the nodes in a belief network be the set $N = \{1, \ldots, n\}$, corresponding to random variables $X_N = \{X_1, \ldots, X_n\}$. Of course, the network is an acyclic directed graph. Each node j has a set of parents $C(j)$, corresponding to the conditioning variables $X_{C(j)}$ for the variable $X_j$. Similarly, $S(k)$ is the set of children of node k corresponding to the variables $X_{S(k)}$ which are conditioned by $X_k$. We assume that the observed evidence is $X_E = x^*_E$, where $E \subset N$, and that we are primarily interested in the posterior marginal probabilities, $P\{X_j \mid x^*_E\}$ for all $j \notin E$. We will use a lower case 'x' to denote the value (instantiation) which variable X assumes. $\leftarrow$ is the assignment operator. $Z \leftarrow Z + A$ means that the new value of Z is set to the sum of the old value of Z and A.

A simple formula underlies the type of Monte Carlo algorithms we are considering. For any given sample $\hat{x}$ selected from the joint distribution of $X_N$, we assign a score, $Z$, equal to the probability of $\hat{x}$ divided by the probability of selecting $\hat{x}$:

$$Z(x[k] | x_E^*) = \frac{\prod_{i=1}^{N} P\{X_i = \hat{x}_i[k] | x_{C(i)}\}}{\prod_{i=1}^{N} P\{\text{selecting } X_i = \hat{x}_i[k] | x_{C(i)}\}}$$

where $\hat{x}[k]$ is the sample made on the $k^{th}$ trial. The probability of selecting x is usually different from the probability of x in the original distribution.

This score is recorded for each instantiation of each unobserved variable,

$$Z(X_j = x_j[k]) = \begin{cases} Z(x[k] | x_E^*), & \text{if } x_j[k] = \hat{x}_j[k] \\ 0, & \text{otherwise.} \end{cases}$$

In most cases, this score is simply accumulated over samples,

$$Z(X_j = x_j) \leftarrow Z(X_j = x_j) + Z(X_j = x_j[k]).$$

Afterwards, the posterior marginal probabilities are estimated by $\hat{P}$, which is derived by normalizing the scores over all possible instantiations of each variable,

$$P\{X_j = x_j | x_E^*\} \approx \hat{P}\{X_j = x_j | x_E^*\}$$
$$\propto Z(X_j = x_j).$$

The simplest example of this type of algorithm is **Logic Sampling** [Henrion 1988]. Each sample x is selected by simulating a value for every variable in the model in graphical order. At the time each variable $X_j$ is simulated, the values of its conditioning variables have already been simulated, $X_{C(j)} = x_{C(j)}$, so $x_j$ is simply selected with probability $P\{x_j | X_{C(j)} = \hat{x}_{C(j)}\}$, given in $X_j$'s conditional probability distribution. Thus the probability of selecting x is given by

$$P\{\text{selecting } x | x_E^*\} = P\{\text{selecting } x\} = P\{x\}$$
$$= \prod_{k \in N} P\{X_k = x_k | x_{C(k)}\}$$

independent of the observed evidence. Of course, we can only count the sample of the evidence generated which corresponds to our observations, so

$$P\{x | x_E^*\} \propto I\{X_E = x_E^*\} \cdot P\{\text{selecting } x\}$$

$$= I\{X_E = x_E^*\} \cdot \prod_{k \in N} P\{X_k = x_k | x_{C(k)}\}$$

and the sample score is
$$Z(x | x_E^*) = I\{X_E = x_E^*\},$$
where

$$I(A) = \begin{cases} 1, & \text{if A is true} \\ 0, & \text{otherwise.} \end{cases}$$

Our **Basic Algorithm** is a minor variation of Logic Sampling [see also Fung and Chang 1989]. A given sample under Logic Sampling is discarded (scored with zeros) whenever $X_E \neq x_E^*$, which happens with probability $1 - P\{x_E^*\}$, our prior probability for the evidence. In the Basic Algorithm, our sample is constrained to correspond to our observation, $X_E = x_E^*$ and any successors to the evidence nodes E in the graph are simulated conditioned on the values observed. Therefore the probability of selecting x is

$$P\{\text{selecting } x | x_E^*\} = \prod_{k \notin E} P\{x_k | x_{C(k)}\}.$$

Now
$$P\{x | x_E^*\} \propto I\{X_E = x_E^*\} \cdot \prod_{k \in N} P\{x_k | x_{C(k)}\}$$
$$= \prod_{k \in N} P\{x_k | x_{C(k)}\},$$

and the sample score is
$$Z(x | x_E^*) = \prod_{k \in E} P\{x_k | x_{C(k)}\}$$

This formula can be clearly recognized as the likelihood function for the unobserved variables given the evidence. Unlike Logic Sampling, the Basic Algorithm will rarely discard a case (assuming nonzero conditional probabilities for $X_E$) even when the observed evidence might have been unlikely beforehand.

The most effective modification of the Basic Algorithm seems to be the **Basic Algorithm with Markov Blanket Scoring**. This modification, suggested in Pearl [1987a] for his stochastic simulation, scores all possible values for a variable at the same time, reducing the variance of the score. Although $x_j$ is the simulated value for variable $X_j$, that value is ignored when scoring $X_j$ and instead scores are generated for all possible values $y_j$. First, a weighting factor is computed for each possible value by temporarily setting $x_j$ to $y_j$,

$$w(y_j) = \prod_{k \in N} P\{x_k | x_{C(k)}\}$$
$$\propto P\{y_j | x_{C(j)}\} \cdot \prod_{k \in S(j)} P\{x_k | y_j, x_{C(k)\setminus\{j\}}\}.$$

After normalizing so that $\sum_{y_j} w(y_j) = 1$,

$$Z_j(y_j) \leftarrow Z_j(y_j) + w(y_j) \cdot Z(x \mid x^*_E)$$

In other words, we are scoring all of the possible states for the variable with a function proportional to probability of the state given its Markov Blanket. Note that this enhancement is applied individually to each variable, so that it can be used selectively, for those variables for which the additional accuracy warrants the additional computational effort.

A common method for improving Monte Carlo approaches is to use an revised "importance" distribution, P', for sampling as an approximation to the posterior distribution (see [Rubinstein 1981] for an explanation of importance sampling). This can be easily applied to the Basic Algorithm, to yield the **Importance Algorithm** (also suggested in [Henrion 1988]). In this case,

$$P\{ \text{selecting } x \mid x^*_E \} = \prod_{k \notin E} P'\{ x_k \mid x_{C(k)} \},$$

so the sample score is

$$Z(x \mid x^*_E) = \prod_{k \in N} P(x_k \mid x_{C(k)}) \bigg/ \prod_{k \notin E} P'(x_k \mid x_{C(k)}).$$

The importance distribution can be generated in many ways, with only two restrictions:
    (1) it cannot be based on the same samples it is used to score; and
    (2) it must be able to sample all possible values, e.g.
$P'\{x_j \mid x^*_E\} = 0$ only if $P\{x_j \mid x^*_E\} = 0$.

In our tests we have considered two different importance distributions: **Self-Importance** and **Heuristic-Importance**. In general, these could be combined with each other and any other heuristics, including rule-based approaches. Note that on problems with no experimental evidence, these algorithms simply become the Basic Algorithm, although in a more complex implementation.

The Self-Importance Algorithm updates its importance distribution using the scores generated in the algorithm,

$$P'_{NEW}\{x_j \mid x_{C(j)}\}$$
$$\propto P'_{OLD}\{x_j \mid x_{C(j)}\} + Z(x \mid x^*_E).$$

Since the renormalization of the P' distribution is a tedious operation, the algorithm tested here performs this update infrequently (every hundred iterations).

The Heuristic Importance Algorithm performs a modified version of the Pearl [1986a] singly connected evidence propagation algorithm to compute likelihood functions, $\lambda(x_j)$, for each of the unobserved variables. Since the network is not in general (or in our tests) singly connected and the likelihood functions are one-dimensional, $\lambda(x_j)$ can be a poor approximation to the the likelihood function. Nonetheless, it appears that $\lambda(x_j) = 0$ only if the exact likelihood is zero as well, and the importance distribution is given by

$$P'\{x_j \mid x_{C(j)}\} \propto \lambda(x_j) \cdot P\{x_j \mid x_{C(j)}\}.$$

## 3. Test Results

A number of tests were performed on the algorithms described in Section 2. Comparisons were made with Logic Sampling [Henrion 1988] and four simulation algorithms based on Stochastic Simulation [Pearl 1987a].

Although the stochastic simulation algorithms are initialized just like the Basic Algorithm, they operate differently thereafter. At each iteration, a single variable is selected randomly for "un-instantiation." It is then re-instantiated to a new value with the probabilities given by the Markov Blanket. Every time a variable is re-instantiated in the **Pearl Algorithm** its state is scored. In the **Pearl with Markov Blanket Scoring**, the Markov blanket probabilities are used to score the re-instantiated variable. A variation on this technique, proposed by Chavez [1989] involves periodic, independent restarts of the Pearl algorithm. Variable states are scored only before restarts, instead of at each re-instantiation. This is called the **Chavez Algorithm**. Finally, if that scoring is done with the Markov blanket probabilities then we obtain the **Chavez with Markov Blanket Scoring**. There are several technical notes about the way these procedures were implemented:

    1. In order to facilitate comparison of the algorithms, we have attempted to keep the number of instantiations per trial constant across all of the algorithms. An instantiation is the assignment of a value to a variable during simulation. Each of the algorithms instantiates a different number of variables per iteration (an iteration is a single "natural" pass of an algorithm). The Basic Algorithm (and variations) instantiates every unobserved variable in the graph during a single iteration. The Pearl and Chavez algorithms instantiate only one variable with each of their iterations. In order to keep the number of instantiations in a test constant for all algorithms, Pearl and Chavez were given more iterations. For example, if the test problem has four unobserved nodes and the Basic Algorithm is given 100 iterations, then the Pearl and Chavez algorithms are given 400 iterations apiece.

    2. The Chavez algorithm is really designed

for a substantially larger number of iterations than we could perform in our tests, so we do not believe that it received a representative showing. Our implementation performs ten restarts on every trial.

We ran each algorithm on our problems with trials of 250 and 1000 iterations. Because of the tremendous variation between multiple trials of the same algorithm on the same problem with such few iterations, we repeated each experiment 25 times and report our mean observations over those 25 trials. At the end of each trial, an error was determined using the formula [Fertig and Mann 1980].

$$\left[ \frac{1}{|N \setminus E|} \sum_{j \notin E} \frac{(\hat{p}_j - p_j)^2}{p_j(1-p_j)} \right]^{1/2}$$

where $\hat{p}_j$ is the computed value for the marginal probabilities and $p_j$ is the exact value. In addition to the mean error, we report the mean time in seconds in ExperCommonLISP on a Macintosh II, the standard deviation of the error, and the product of squared mean error and mean time, which appears to be fairly invariant for some algorithms as we increase the number of iterations.

There were two problems we analyzed for which results are presented here. The first is a cancer example, shown in Figure 1, which first appeared in [Cooper 1984] and has become something of a standard test problem of a multiply-connected network. The probabilities and evidence we used for this problem are summarized in [Pearl 1987b]. It was tested both without any experimental evidence and with the observation of severe headaches without coma. The results of our tests are summarized in Table 1.

We also created a simple problem with some deterministic logical nodes to see how well our algorithms could perform. Although none of the nonzero probabilities were extreme, there are some logical nodes (OR's and AND's) in the network, shown in Figure 2. For problems of this sort, Pearl and Chavez are no longer theoretically guaranteed to converge to the correct answer, since the embedded Markov chain is no longer ergodic. We could have devised problems which were even more (or less) pathological for those algorithms, but our real concern was testing our new algorithms against a realistic problem. The conditional probabilities in our model are:

P{ A } = P{ B } = .9;
P{ C | B } = .9; P{ C | ¬B} = .1
P{ D | B, C } = .9; P{ D | B, ¬C } = .8;
P{ D | ¬B, C } = .2; P{ D | ¬B, ¬C} = .1;
P{ E | AND } = .9; P{ E | ¬AND } = .1 .

We tested this problem with no experimental evidence and with evidence that E is true, and those results are summarized in Table 2.

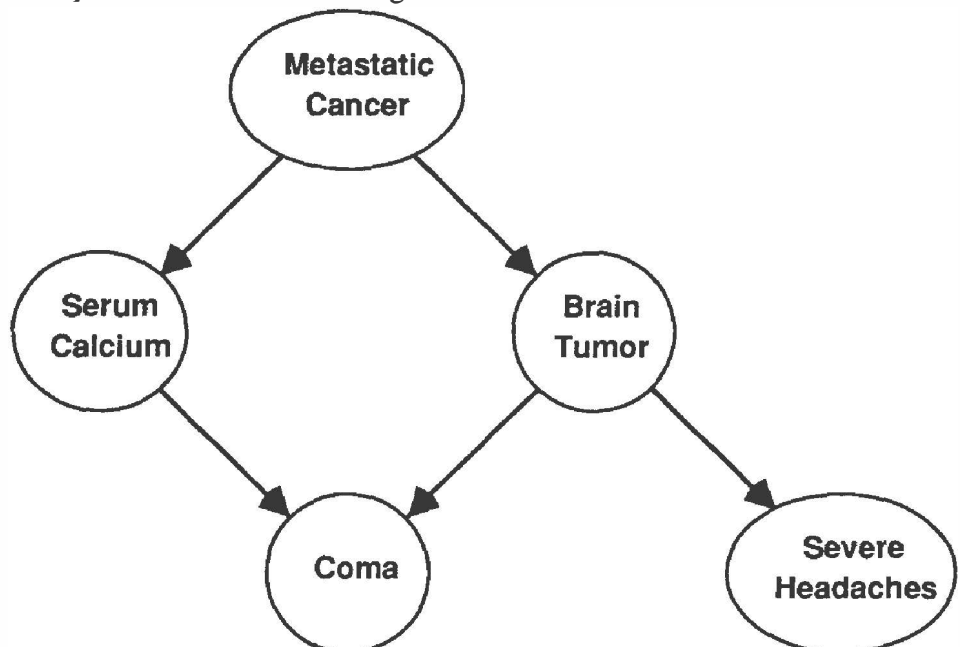

Figure 1. Multiply Connected Belief Network from Cooper [1984].

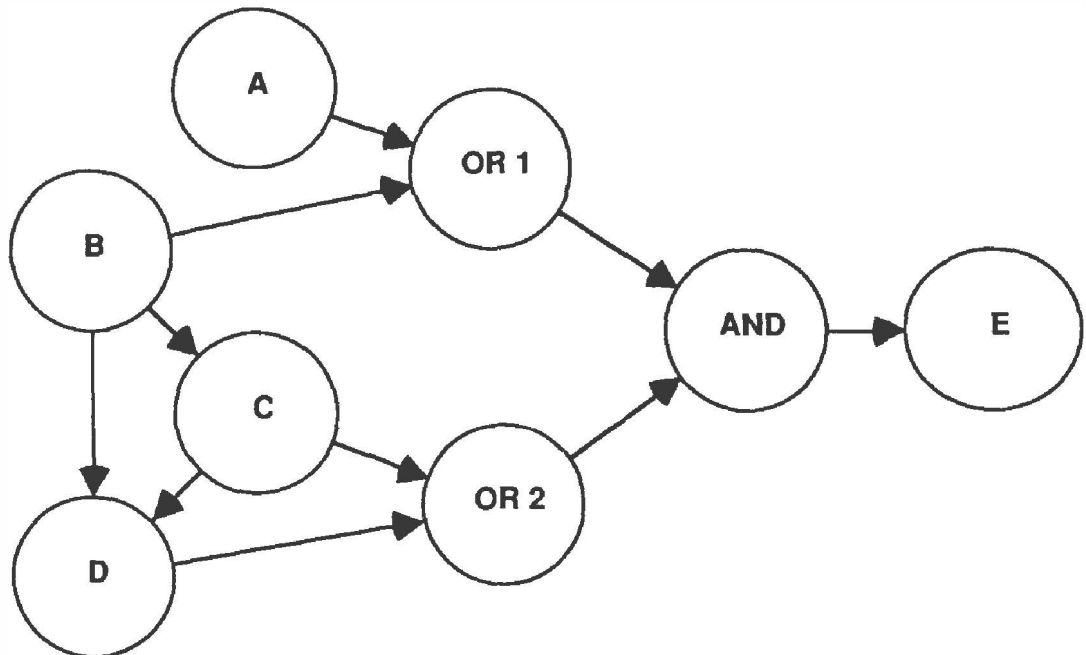

Figure 2. Multiply Connected Network with Some Logical Determinacy.

## 4. Conclusions

First, let us consider the results of the race. The Basic Algorithm and Markov Blanket Algorithm were winners for each case considered and overall. Among Markov chain algorithms, Pearl with Markov Blanket Algorithm was the best, although it has the problems one would expect when there are deterministic (or near deterministic) functions present. The Self-Importance and Heuristic-Importance functions performed poorly compared to the other new algorithms. This would suggest that the successful use of such techniques would depend on better heuristics for the importance weighting.

There are several ways that the suggested approached can be integrated into a general algorithm. The Markov Blanket modification to the Basic Algorithm can be applied selectively on a node-by-node basis. Thus it can be decided whether a particular node merits the additional computations either due to its utility to the decision maker or because of statistics generated so far during the run, if, for example, it appears to have an extreme distribution.

Other modifications to the Basic Algorithm (and hence the others) allow additional resources to be used to refine the computation at a particular node. The evidence likelihood used to score any node j does not necessarily depend on every evidence node, but rather on $E \cap N_\pi(j \mid E)$, as discussed in [Shachter 1988]. Using this $N_\pi$ operator, the nodes in the network can be partitioned by their relevant evidence, or processed separately for the most variance reduction. In fact, if posterior marginals are only needed for the nodes in $S \subset N$, then we only need to simulate the variables $X_{N_\pi(S \mid E)}$, with evidence likelihood for $E \cap N_\pi(S \mid E)$. Again, these enhancements can be applied adaptively to improve the accuracy of the simulation.

The algorithm is suitable for parallel processing architectures. Each processor can run its own copy of the simulation and all copies of the simulation can be combined in the end by adding the scores. Of course, the simulation can interrupted at any time and will return the best quality answer thus far along with an estimate of the standard error based on the sample variance of the generating process.

Finally, there is one type of problem structure for which the Basic Algorithm would be poor: when the likelihood product varies greatly between iterations, especially when it is occasionally much larger than usual. The result of this behavior is that most of the iterations are ignored, and this is the problem with Logic Sampling. There are two cures in this situation. One is to develop a good importance weighting scheme so that most iterations are fairly consistent with the evidence. This

|  | Logic Sampling | Basic Algorithm | Markov Blanket | Self Importance | Heuristic Importance | Pearl | Pearl with M. Blanket | Chavez | Chavez with M. Blanket |
|---|---|---|---|---|---|---|---|---|---|
| **Cooper Example with no Evidence** | | | | | | | | | |
| **250 Iterations, 25 trials** | | | | | | | | | |
| mean error | 0.058 | 0.056 | 0.022 | 0.062 | 0.063 | 0.122 | 0.059 | 0.308 | 0.182 |
| std. deviation error | 0.021 | 0.020 | 0.014 | 0.030 | 0.022 | 0.059 | 0.036 | 0.098 | 0.094 |
| mean time (sec) | 8.288 | 8.369 | 24.045 | 17.628 | 15.207 | 23.357 | 25.378 | 23.344 | 24.044 |
| error$^2$ time | 0.028 | 0.026 | 0.011 | 0.069 | 0.060 | 0.347 | 0.087 | 2.213 | 0.796 |
| **1000 Iterations, 25 trials** | | | | | | | | | |
| mean error | 0.031 | 0.029 | 0.009 | 0.029 | 0.028 | 0.071 | 0.031 | 0.292 | 0.188 |
| std. deviation error | 0.016 | 0.008 | 0.005 | 0.014 | 0.012 | 0.043 | 0.021 | 0.102 | 0.085 |
| mean time (sec) | 33.021 | 34.639 | 99.760 | 72.670 | 61.983 | 96.983 | 105.925 | 95.482 | 96.758 |
| error$^2$ time | 0.032 | 0.030 | 0.008 | 0.060 | 0.049 | 0.487 | 0.104 | 8.128 | 3.428 |
| **Cooper Example with Published Evidence** | | | | | | | | | |
| **250 Iterations, 25 trials** | | | | | | | | | |
| mean error | 0.094 | 0.049 | 0.015 | 0.039 | 0.066 | 0.078 | 0.035 | 0.281 | 0.121 |
| std. deviation error | 0.039 | 0.025 | 0.006 | 0.017 | 0.043 | 0.045 | 0.022 | 0.082 | 0.065 |
| mean time (sec) | 7.566 | 6.835 | 21.425 | 12.549 | 11.190 | 16.564 | 17.662 | 16.515 | 16.957 |
| error$^2$ time | 0.067 | 0.017 | 0.005 | 0.019 | 0.048 | 0.101 | 0.021 | 1.307 | 0.248 |
| **1000 Iterations, 25 trials** | | | | | | | | | |
| mean error | 0.042 | 0.018 | 0.009 | 0.023 | 0.034 | 0.046 | 0.014 | 0.282 | 0.107 |
| std. deviation error | 0.020 | 0.008 | 0.005 | 0.009 | 0.018 | 0.024 | 0.008 | 0.151 | 0.048 |
| mean time (sec) | 30.473 | 28.113 | 88.692 | 51.383 | 45.369 | 68.218 | 73.247 | 67.644 | 68.083 |
| error$^2$ time | 0.053 | 0.009 | 0.007 | 0.027 | 0.054 | 0.143 | 0.015 | 5.380 | 0.778 |

Table 1. Comparisons for Cooper Example Shown in Figure 1.

*Note: Some key explanations of the data in this table are presented in the text.*

|  | Logic Sampling | Basic Algorithm | Markov Blanket | Self Importance | Heuristic Importance | Pearl | Pearl with M. Blanket | Chavez | Chavez with M. Blanket |
|---|---|---|---|---|---|---|---|---|---|
| **Deterministic Example with no Evidence** | | | | | | | | | |
| **250 Iterations, 25 trials** | | | | | | | | | |
| mean error | 0.059 | 0.062 | 0.042 | 0.059 | 0.058 | 0.543 | 0.690 | 0.277 | 0.207 |
| std. deviation error | 0.020 | 0.024 | 0.023 | 0.023 | 0.018 | 0.523 | 0.889 | 0.116 | 0.121 |
| mean time (sec) | 12.453 | 13.649 | 35.478 | 30.145 | 26.572 | 40.022 | 42.714 | 39.837 | 41.176 |
| error$^2$ time | 0.044 | 0.053 | 0.061 | 0.104 | 0.090 | 11.806 | 20.357 | 3.059 | 1.771 |
| **1000 Iterations, 25 trials** | | | | | | | | | |
| mean error | 0.031 | 0.031 | 0.021 | 0.031 | 0.026 | 0.426 | 0.592 | 0.284 | 0.215 |
| std. deviation error | 0.012 | 0.012 | 0.009 | 0.015 | 0.011 | 0.384 | 0.568 | 0.107 | 0.137 |
| mean time (sec) | 50.187 | 54.662 | 140.633 | 120.326 | 104.019 | 158.901 | 171.056 | 157.822 | 159.365 |
| error$^2$ time | 0.049 | 0.051 | 0.062 | 0.119 | 0.072 | 28.803 | 60.032 | 12.721 | 7.372 |
| **Deterministic Example with Evidence "true"** | | | | | | | | | |
| **250 Iterations, 25 trials** | | | | | | | | | |
| mean error | 0.063 | 0.043 | 0.020 | 0.046 | 0.053 | 0.271 | 0.945 | 1.001 | 0.489 |
| std. deviation error | 0.024 | 0.014 | 0.007 | 0.013 | 0.020 | 0.853 | 1.742 | 0.591 | 0.466 |
| mean time (sec) | 12.499 | 14.383 | 47.505 | 27.534 | 24.237 | 35.879 | 38.540 | 36.239 | 37.335 |
| error$^2$ time | 0.049 | 0.027 | 0.019 | 0.057 | 0.069 | 2.628 | 34.395 | 36.298 | 8.937 |
| **1000 Iterations, 25 trials** | | | | | | | | | |
| mean error | 0.036 | 0.023 | 0.011 | 0.021 | 0.025 | 0.433 | 0.873 | 0.849 | 0.673 |
| std. deviation error | 0.012 | 0.006 | 0.004 | 0.008 | 0.007 | 1.182 | 2.443 | 0.427 | 0.464 |
| mean time (sec) | 49.661 | 57.183 | 188.159 | 109.617 | 95.017 | 143.859 | 153.585 | 143.396 | 144.361 |
| error$^2$ time | 0.064 | 0.030 | 0.021 | 0.051 | 0.060 | 27.017 | 117.068 | 103.463 | 65.375 |

**Table 2. Comparisons for Deterministic Example Shown in Figure 2.**

*Note: Some key explanations of the data in this table are presented in the text.*

weighting might be derived through interaction with an expert; for example, the user of a system to support medical diagnosis might suggest likely diseases given the evidence set. This information may be used to modify the importance distribution to concentrate the 'probabilistic mass' of the samples in areas where the expert feels that the joint probability for the solution is high. A knowledge based system could be used in the same way to suggest a good importance distribution. The other cure is to reverse the arcs to those evidence nodes most responsible for the variation in likelihood. Although this operation can be expensive to perform, since it increases the complexity of the diagram, it will significantly reduce the variation in likelihoods between iterations [Fung and Chang 1989]. If carried to its ultimate extreme, in which evidence variables are only conditioned by other evidence variables, the likelihood becomes a constant as in the case of no observed experimental evidence [Chin and Cooper 1987].

One important deficiency of this work is the lack of a theoretical bound on the convergence rate of the algorithm. The results of [Henrion 1988] are applicable for the **Basic Algorithm** with no observed experimental evidence, but we have not found an upper bound on convergence rate for the other algorithms or for the Basic Algorithm with observed evidence.

In summary, the Basic and Markov Blanket Algorithms appear to provide competitive performance. Although other exact and approximate procedures are superior for problems with special structure, our proposed algorithms are simple and robust for a wide variaty of probabilistic inference problems.

## 5. Acknowledgments

We benefited greatly from discussions with Bob Fung, Martin Chavez, Greg Cooper, and Max Henrion.